\documentclass[conference]{IEEEtran}
\IEEEoverridecommandlockouts
\usepackage{cite}
\usepackage{amsmath,amssymb,amsfonts}
\usepackage{algorithmic}
\usepackage{graphicx}
\usepackage{textcomp}
\usepackage{xcolor}
\usepackage{color}
\usepackage{bm}
\usepackage[T1]{fontenc}
\usepackage[utf8]{inputenc}
\usepackage{authblk}
\def\BibTeX{{\rm B\kern-.05em{\sc i\kern-.025em b}\kern-.08em
    T\kern-.1667em\lower.7ex\hbox{E}\kern-.125emX}}

\begin{document}
\title{CMR motion artifact correction using generative adversarial nets\\}
\author[*]{Yunxuan Zhang}
\author[**]{Weiliang Zhang}
\author[**]{Qinyan Zhang}
\author[*]{Jijiang Yang}
\author[***]{Xiuyu Chen}
\author[***]{Shihua Zhao}
\affil[*]{Research Institute of Information Technology.Tsinghua University,Beijing,China}
\affil[**]{Automation School.Beijing University of Posts and Telecommunications,Beijing,China}
\affil[**]{MRI Center.Fuwai Hospital.Chinese Academy of Medical Science,Beijing,China}
\affil{Corresponding author: yangjijiang@tsinghua.edu.cn}
\maketitle

\begin{abstract}
Cardiovascular Magnetic Resonance (CMR) plays an important role in the diagnoses and treatment of cardiovascular diseases while motion artifacts which are formed during the scanning process of CMR seriously affects doctors to find the exact focus.
The current correction methods mainly focus on the K-space which is a grid of raw data obtained from the MR signal directly and then transfer to CMR image by inverse Fourier transform.
They are neither effective nor efficient and can not be utilized in clinic.
In this paper, we propose a novel approach for CMR motion artifact correction using deep learning. Specially, we use deep residual network (ResNet) as net framework and train our model in adversarial manner.
Our approach is motivated by the connection between image motion blur and CMR motion artifact, so we can transfer methods from motion-deblur where deep learning has made great progress to CMR motion-correction successfully.
To evaluate motion artifact correction methods, we propose a novel algorithm on how edge detection results are improved by deblurred algorithm.
Boosted by deep learning and adversarial training algorithm, our model is trainable in an end-to-end manner, can be tested in real-time and achieves the state-of-art results for CMR correction.
\end{abstract}


\section{Introduction}

\begin{figure}[htbp]
	\centerline{\includegraphics[width=6cm, height=10cm]{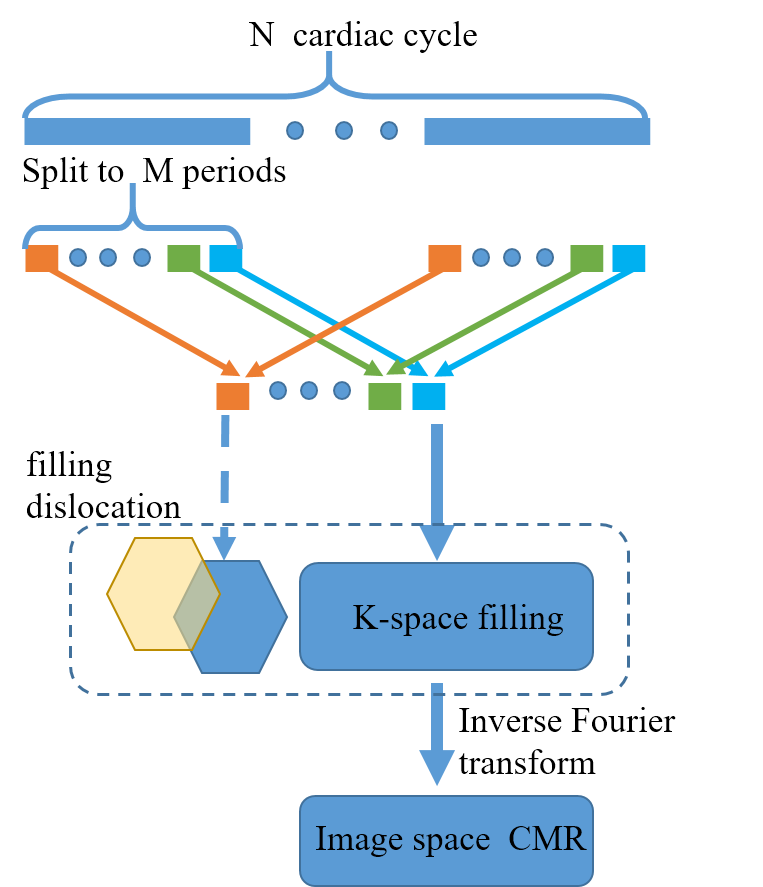}}
	\caption{Illumination of the CMR motion artifact process. The process of CMR generation is shown by hard lines.  If there is irregular fibrillation during scanning, the information from the same periods will be filled in different places which cause motion artifact, shown by the dash line.}
	\label{motion artifact}
\end{figure}

Cardiovascular Magnetic Resonance (CMR) is widely used in clinic for diagnoses and treatment of cardiovascular disease.
The motion artifacts generally existing in CMR will degrade the quality of images, especially blur the boundary between the inner lining and the outer lining of heart and affect doctors to find the precise focus location in the clinic.
Motion artifact in CMR is mainly caused by patients' atrial fibrillation during the procedure of scanning which will induce phase distortion in the collected signals then there will be motion artifact in the reconstructed image after inverse Fourier Transform, shown in  Fig.~\ref{motion artifact}. The current machine can't complete a CMR within a cardiac cycle so we have to combine information from N cardiac cycles, N usually is 20. In particular, we split each cardiac cycle to M periods and combine the same period from N cycles to the corresponding period of K-space which is a kind of phase space and values correspond to spatial frequencies of MR image.

\begin{figure}[htbp]
    \centerline{\includegraphics[width=9cm, height=11cm]{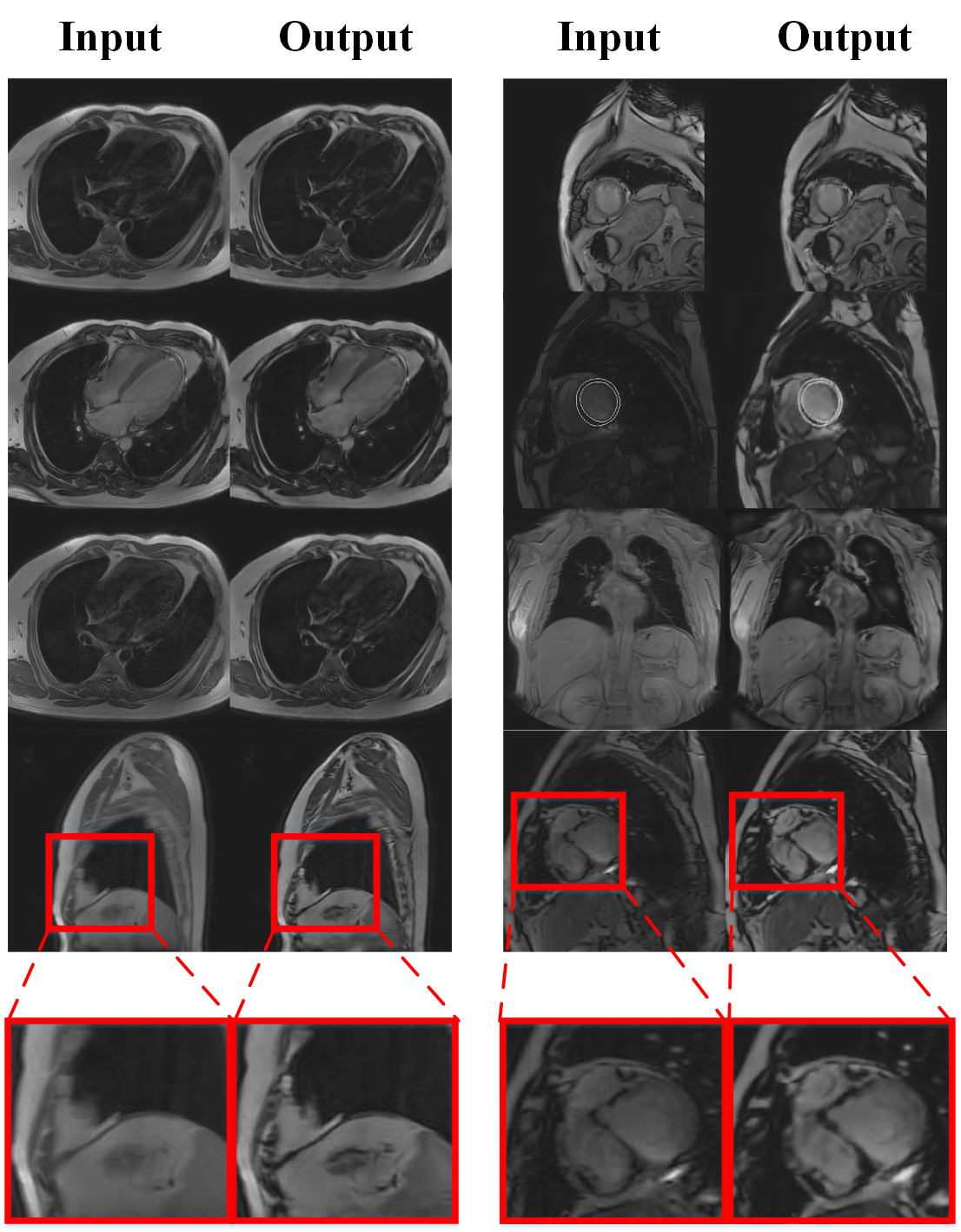}}
	\caption{Results of CMR correction model for different parts.
	The odd columns are inputs with motion artifact while the even columns are corresponding correction results.
	The details are highlighted by colored rects.
	It is obvious that correction results sharper than inputs.}
	\label{Result}
\end{figure}

CMR motion artifact correction has received many attention in recent years, but most of the current correction methods \cite{Hirokawa2008MRI, Junping2012Discrimination, Nyberg2012Comparison, Tan20113D} are applied in K-space and rely on large effort and delicacy design on parameter fitting which is time-cost and unfeasible in practice.
Considering their complexity, few of them are open-sourced. So far, there is no effective or complete system for CMR motion artifact correction.

Our work is devoted to correct motion artifact in CMR using deep learning. Specially we use residual network \cite{he2016deep} as principle net framework to reconstruct sharp image taking blur image as input and train our model in the adversarial manner \cite{goodfellow2014generative} to preserve more details which can make output look like more realistic, as shown in Fig.~\ref{Result}.

%\xk{write motivation first, then the detail formulations.}
Our method is motivated by the connection between image motion blur and CMR motion artifact. With the connection, we can transfer significant motion deblur techniques from computer vision domain to CMR correction successfully.
Indeed, the motion deblurs in CV domain and motion artifact correction in CMR have essential similarity in principle.

Equation \eqref{principle} is the definition of motion blur,
\begin{equation}
L_{blur} = Kernel * L_{sharp} + \epsilon,
\label{principle}
\end{equation}

where $L_{blur}$ and $L_{sharp}$ are latent space of blur image and sharp image respectively, $*$ refers to convolution operation, $Kernel$ refers to blur kernel and $\epsilon \sim N(0,\sigma^2)$ for a specific $\sigma$.

 With regard to CMR motion artifact, K-space can be regarded as a particular image latent space, atrial fibrillation is the source of blur kernel.
 This connection provides us theoretical support for transferring deblur method from Computer Vision to CMR motion artifact correction so that we can leverage the advantages of ResNet in image processing.

We present the following three principal contributions.

First, We find the connection between the theories of motion-deblur in CV domain and motion artifact correction in CMR so we can transfer methods and take advantages of Convolution Neural Network(CNN) in image processing.

Second, we develop a  convolutional neural network architecture for CMR motion artifact correction (we call it CMR Motion artifact Correction Network, short for \textbf{CMCN}) that is trainable in an end-to-end manner taking the blur-sharp image pairs as input and supervision.
The framework is AutoEncoder likely and we apply residual network to make the network deeper which is good for parameter learning and apply adversarial generate network to enhance the generalization of system and reality of the restored output. More details about net framework can be find in \ref{net_arc}.
Overall, the framework is easy to implement and open-sourced.

Specifically, Adversarial training framework proposed by generative adversarial networks shows promise in both image generation and image-to-image translation, and deep residual networks provide valid framework for many computer vision tasks.
We combine the advantages of adversarial training and deep residual networks and the experiments show that our model achieve the state-of-the-art correction for CMR motion artifact.

Finally, we propose a novel method for evaluation correction algorithms based on how they improve image edge detection results.
We assume that sharper CMR has clearer organ boundary which leads to accurate diagnosis and treatment in real-clinic problems.
So our approach focuses on evaluating the sharpness of the organ boundary in generated images and it can reveal the quality of correction models better than standard PSNR metric.
Especially, we choose the edge connectivity as evaluate metric which is typical assessment metric for edge detection, in general, when using the same edge detection algorithm in different images, the metric is lower on sharp images while it's harder to detect the edge in the image that has motion artifacts.
The details about evaluate metrics can be find in \ref{metric_result}.

\section{Related work}

\subsection{CMR motion artifact correction \& Image motion deblur}
CMR motion artifacts are very common in cardiac surgery medical images. It will affect the doctor's judgment of the patient's condition, so, it is very necessary to remove motion artifacts. YH Tseng\cite{Yen1994postprocessing} presented a new post-processing algorithm to deal with more general motion artifacts. This algorithm corrects blurry images by constantly iterating through the knowledge of the image. Wu Chunli\cite{Chunli2015TranslationalMotion} presented an image correction algorithm which combined Fourier projection algorithm and genetic algorithm for handling CMR translational motion artifacts. This method has higher image clarity and faster imaging speed at that time. Huang Min\cite{HUANGMin2013minimumentropyconstraint} improved the MRI motion artifact correction method based on minimum entropy constraint, which improved the correction effect. The above methods are traditional methods, then the effect of improvement is limited.

Image motion deblur model is defined in \eqref{principle}. The early algorithms are based on an image with a clear blur kernel K. Most of them rely on classics Lucy-Richardson algorithm. After many iterations of the algorithm, the image can have a good deblurring effect. However, most of the blurred images in daily life do not know the blur kernel K. In recent years, with the extensive application of deep learning in the image domain, CNN has been widely used in the study of image deblurring and has achieved good results. Ankit Gupta et al\cite{Gupta2010single} presented a completely new method for dealing with the blur of a single image to estimate spatial non-uniform blur produced by camera shake. In this methods, the camera motion is represented as a Motion Density Function (MDF) and based on it, a novel deblurring method is proposed. Li Xu et al\cite{Xu2013Unnatural} presented a generalized and mathematically sound L0 sparse expression, they also presented a new effective method to deal with motion deblurring. Jian Sun et al\cite{Sun2015Learning} used convolutional neural networks to predict the probability distribution of motion blur and used the Markov random field model to infer non-uniform motion blur fields and got good results. Deep Learning Technology Provides an unprecedented research idea and method for image deblurring research.
\subsection{Generative adversarial networks}
The idea of generating adversarial networks comes from the two-player game, this kind of idea belongs to a kind of game theory. The idea of GAN was first proposed by Goodfellow et al \cite{goodfellow2014generative}. GAN contains two models, the discriminator and the generator. The generator set noise as input, and generate a sample as output. The discriminator is used to distinguish between the real sample and the generated sample. The purpose of the generator is to generate a real model as much as possible and the purpose of the discriminator is to distinguish the generation sample and the real sample as accurately as possible. From the theoretical perspective, the game between the generator G and discriminator D is the minimax objective.

The \eqref{gan loss} corresponds to two optimization processes, max D and min G. What we need to do is to make G produce the same data as possible in the data set, so we need to minimize the error of the generated model. $D_{\theta_D}(I_{sharp})$ is the output of the discriminant model and represents the probability that the input x is real data. Our goal is to make the output of the discriminant model output $D_{\theta_D}(I_{sharp})$ as close to 1 as possible, so we need to maximize D. Least Squares GANs(lsgans) \cite{Xudong2017lsgans} was introduced by Xudong Mao et al. This method aims at the improvement of the two defects of the standard GAN generated images are not high quality and the instability of the training process. The improved method is to change the objective function of GAN from the cross-entropy loss to the least squares loss. WGAN  \cite{Martin2017wgan} use Wasserstein distance instead of KL divergence.  The Wasserstein distance is smooth and can provide meaningful gradients, which will solve the issue of gradient disappearance. GAN is a very significant research direction in the domain of deep learning and it can improve the generalization performance of the model.
\subsection{Evaluation metric for Motion artifact correction}
The purpose of our paper is to use blurred images with motion artifacts to generate sharp images. So we need to use some evaluation criteria to determine the similarity between images. In our paper, we use Structural Similarity(SSIM), Peak Signal to Noise Ratio(PSNR), edge connectivity as our evaluation criteria.
SSIM was first introduced by Zhou Wang\cite{ZhouWang2004ssim} et al who come from Laboratory for Image and Video Engineering. SSIM is a measure of the similarity between two images. We can use Python's library to calculate SSIM. The peak signal-to-noise ratio (PSNR) is an objective measure of image distortion or noise level. The greater the PSNR between two images, the more similar the two images are. There is also an evaluation method based on connected components. We calculate 4-connected component number B, 8-connected component number C, the number of image edge points A and the ratio of C/B and C/A. The size of C/B and C/A reflects the degree of edge linear connectivity. We could get better edge effect images when we get a smaller value of C/B and C/A. By studying the principles of the above three evaluation criteria, we think that they are applicable to our CMR images.

\section{Proposed method}
In this section, we first illustrate the network architecture, then we demonstrate the loss functions, finally we introduce our evaluation metric.

\begin{figure}[htbp]
	\centerline{\includegraphics[width=9cm, height=7cm]{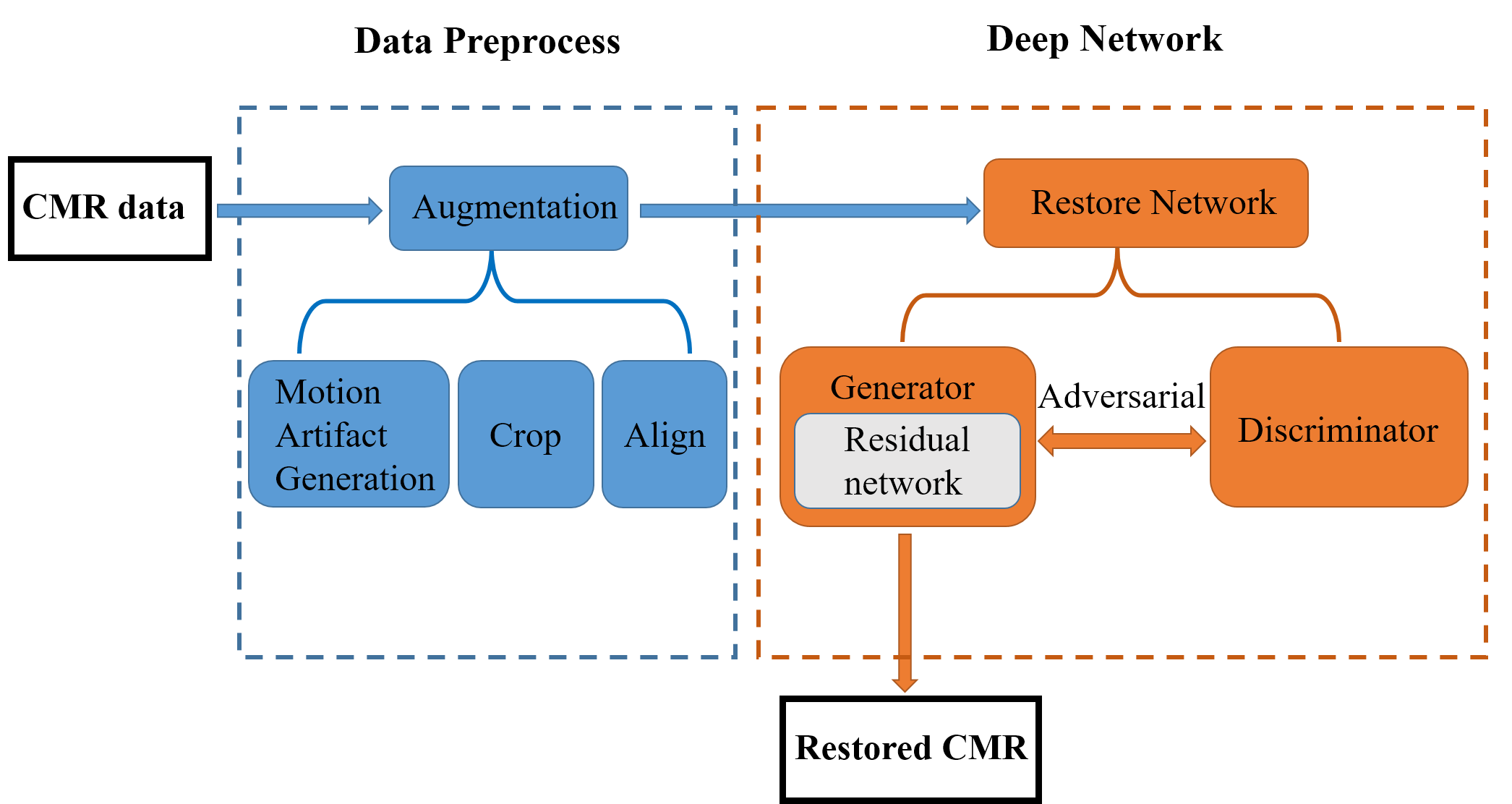}}
	\caption{The pipeline of our CMR motion artifact correction system based on deep learning. The data preprocess including motion artifact generation, crop and align is shown in blue, and the deep network is shown in orange.}
	\label{system}
\end{figure}

Fig.~\ref{system} illustrates the overall framework of our CMR motion artifact correction system. Given an blur CMR as input, the Generator $G_{\theta_G}$ restores a sharper output.

Suppose there are two image domains: $\mathcal{M}$ and $\mathcal{S}$, referring to image domain with and without (w/o) motion artifact respectively.
Analytically, our goal is to restore sharp image $I_S$ given only a motion artifact image $I_M$, with no information about motion trajectory.
In our strategy, we train an Generator $G_{\theta_G}: \mathcal{M}\rightarrow \mathcal{S}$ to correct $I_M$ from domain $\mathcal{M}$ to its corresponding $I_S$ in target domain with sharp texture.
During the training phase, we introduce the discriminator $D_{\theta_D}$ in an adversarial manner to generate more realistic images.

Overall, as discussed above, our correction model has three players, including $G_{\theta_G}$, $D_{\theta_D}$ and $E$, where the first two coorperate in adversarial manner to restore blur CMR and $E$ refers to the Edge net whose output is the edge of the input.
Besides, $\lambda_{gan}$ and $lambda_{edge}$ are loss weights.
In our work, $G_{\theta_G}$ is also supervised by $E$ which help to restore more clear edge of CMR.
The loss of our framework has three terms, as shown in \eqref{loss function}:

\begin{equation}
\mathcal{L} = \mathcal{L}_{content} + \lambda_{gan}\mathcal{L}_{GAN} + \lambda_{edge}\mathcal{L}_{edge}.
\label{loss function}
\end{equation}

To evaluate the quality of CMR motion artifact correction models for clinic purpose, we propose a novel evaluate metric based on the edge detection results.
Especially, we choose the Edge Connectivity as quantitive score.
It can be demonstrated that our evaluate metric more reasonable than the conventional metric such as PSNR in Fig.~\ref{metric_result}.

\subsection{Network architecture} \label{net_arc}
\begin{figure}[htbp]
	\centerline{\includegraphics[width=9cm, height=7.5cm]{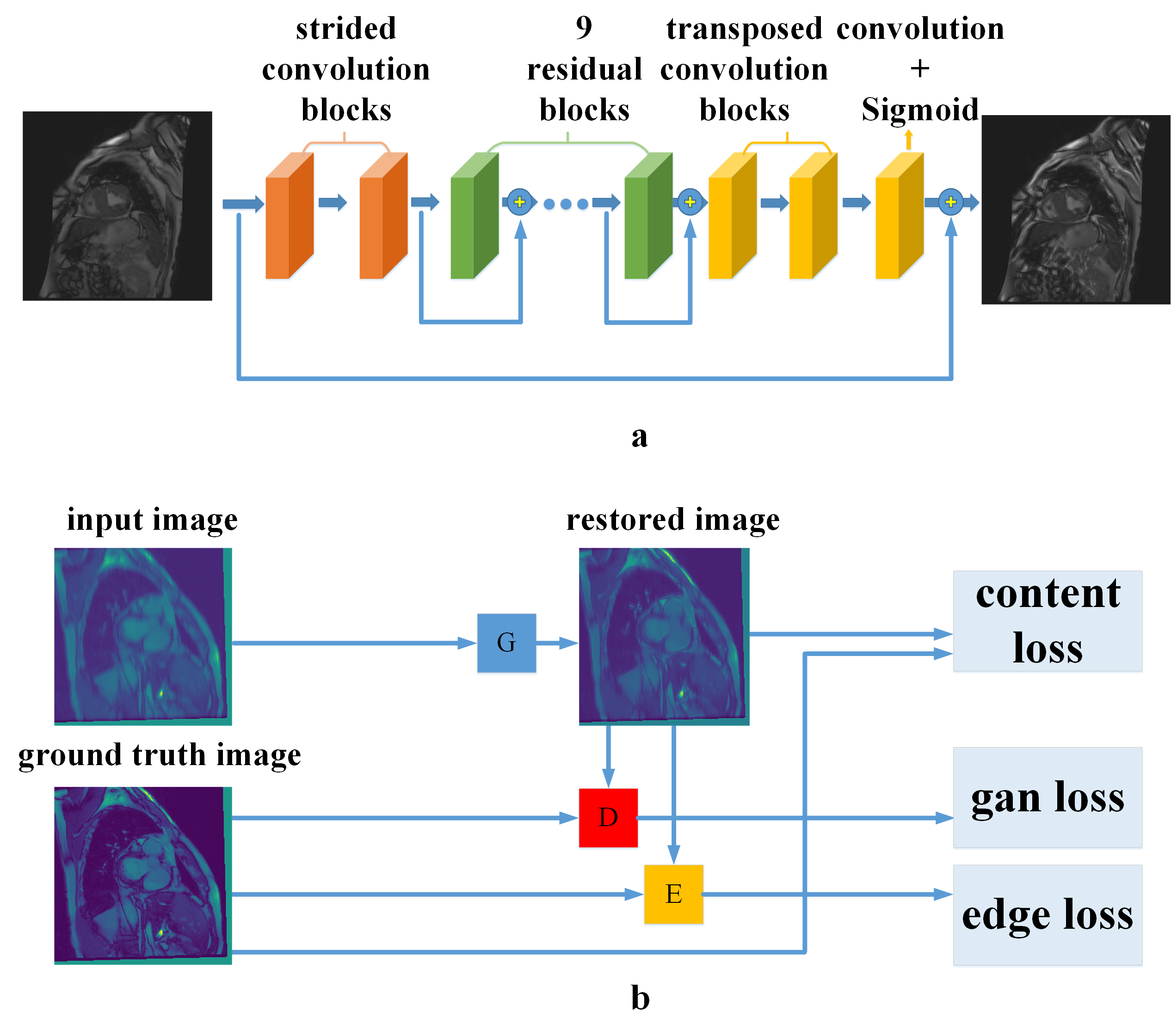}}
	\caption{General architecture of our model. (a) is the architecture of generator network which contains two strided convolution blocks with stride 1/2 , nine residual blocks and two transposed convolution blocks. Each strided convolution block consists of a 3$\times$3 convolution layer, instance normalization layer, and ReLU activation. Each ResBlock consists of a 3$\times$3 convolution layer, instance normalization layer, and ReLU activation. Each transposed convolution blocks consists of a transposed convolution layer, instance normalization layer, and ReLU activation. (b) is the overall training flow, $G$ is generator, $D$ is the discriminator, $E$ is Edge filter.}
	\label{net_archi}
\end{figure}

Our network architecture is shown in Fig.~\ref{net_archi}. We adopt the architectures of our transformer and discriminator from those in \cite{Kupyn2017DeblurGAN}. Specifically, the generative network consists of two stride-2
convolutions, nine residual blocks, and two fractionally-strided convolutions. The discriminative network is implemented with Global GAN \cite{goodfellow2014generative}. Although Patch GAN \cite{zhu2017unpaired} has the advantage of fewer parameters and being able to be adapted to images with arbitrary size, CMR data that has similar background between the image with and w/o motion artifact will confuse the $D_{\theta_D}$. And in that case, global gan can capture the overall feature of the image and neglect the confusing information of background.

\subsection{Loss function}\label{loss section}
Our learning-based framework take an image with motion artifact as input and restore a corresponding sharp image.
It is pix2pix likely, thus we can derive the loss for $G_{\theta_G}$ and $D_{\theta_D}$'s parameter.
First, we employ a very popular loss which is the combination of content loss and GAN loss.
This loss is widely used at recent work, such as Pix2Pix\cite{isola2017image} and plug and play GAN\cite{nguyen2016plug}.
The third term is edge loss, we introduce it to let the $G_{\theta_G}$ learn sharper edge for clinic purpose. The $\lambda_{gan}$ and $\lambda_{edge}$ are the weights to balance losses.
During training phase, we set $\lambda_{gan} = 100$ and $\lambda_{edge} = 100$.

\textbf{Content Loss} is widely used in pix2pix task to constrain the similiar content and construction between output and target.
\textbf{L1} or \textbf{MAE} loss, \textbf{L2} or \textbf{MSE} loss are two classical choices for ``content'' loss function on raw pixels.
As shown in \cite{isola2017image}, \textbf{L1} loss has advantage in convergence.
 So we adopted \textbf{L1} loss, it is defined as following \eqref{content loss} where $I_{blur}$, $I_{restored}$ and $I_{target}$ refer to image with motion artifact, image restored by model and supervision respectively, especially $I_{blur} \in \mathcal{M}$ and $I_{sharp} \in \mathcal{S}$.
 And channel, height, width are the dimensions of image:

\begin{equation}
\label{content loss}
\left\{\begin{array}{l}
\mathcal{L}_{content}(I_{restored}, I_{target}) = mean({l_1,...,l_N}^T) \\[0.3cm]
I_{restored} = G_{\theta_G}(I_{blur}) \\[0.3cm]
l_n = |{I_{restored}}_n - {I_{target}}_n |\\[0.3cm]
N = channel \times height \times width
.
\end{array}\right.
\end{equation}

\textbf{GAN Loss} As though it has been demonstrated in \cite{Kupyn2017DeblurGAN} that patch gan loss perform better than global gan loss, the black background of CMR which lead to the same patch value confuses the discriminator.
So we adopt vanilla GAN loss \cite{goodfellow2014generative} as the critic function .
GAN loss, shown in \eqref{gan loss}, can help $G_{\theta_G}$ generate more realistic details in outputs. Discriminator $D_{\theta_D}$ are trained to distinguish real sharp image from restored image generated by $G_{\theta_G}$, while $G_{\theta_G}$ are trained to generate restored image that confuse the $D_{\theta_D}$.
$G_{\theta_G}$ and $D_{\theta_D}$ are trained in an adversarial manner.

\begin{equation}
\label{gan loss}
\begin{split}
\mathcal{L}_{gan}(D_{\theta_D}, G_{\theta_G}) =
&\min\limits_{G_{\theta_G}}\ \max\limits_{D_{\theta_D}} \mathop{E}\limits_{I_{sharp}\sim I_S}[log{D_{\theta_D}(I_{sharp})}] \\
&+ \mathop{E}\limits_{I_{blur}\sim I_M}  [log(1-D_{\theta_D}(G_{\theta_G}(I_{blur})))]
,
\end{split}
\end{equation}

\textbf{Edge Loss} To let the restored image have more details in the edge.
 Here we add another edge constraint loss, shown in \eqref{edge loss}, directly between the input and output of a $G_{\theta_G}$.
  Here we first obtain edge of the image using edge detection method $E(\cdot)$. Then we use \textbf{L1} loss as criterion function to intensify the edge of output.
  Especially, we use Sobel operator \cite{sobel2014history} as edge detection function.
  In order to let $E(\cdot)$ compatible with our network and derivable, we realize the Sobel operator as a Convolution layer.
  Thus we have our edge constraint loss as shown in \eqref{edge loss}.

\begin{equation}
\label{edge loss}
\left\{\begin{array}{l}
\mathcal{L}_{edge}(I_{restored}, I_{target}) = mean({l_1,...,l_N}^T) \\[0.3cm]
I_{restored} = G_{\theta_G}(I_{blur}) \\[0.3cm]
l_n = |E({(I_{restored}}_n) - E({I_{target}}_n) |\\[0.3cm]
N = channel \times height \times width
.
\end{array}\right.
\end{equation}

\subsection{Evaluation} \label{metric_result}

\begin{figure}[htbp]
	\centerline{\includegraphics[width=9cm, height=4cm]{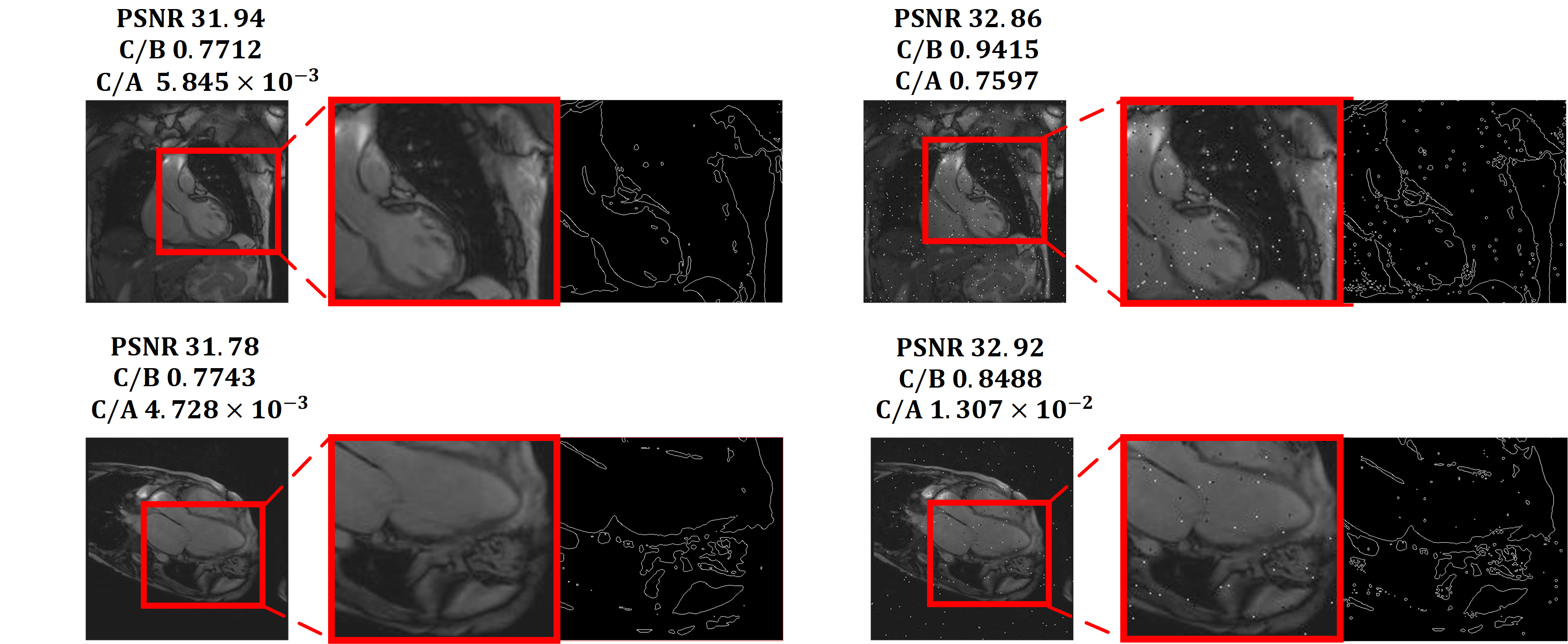}}
	\caption{Some cases with high PSNR but low quality in fact, in contrast, score for edge detection is low. There are four parts and each contains CMR image, details highlighted by red and its corresponding edge. In each row, the CMR in left and right part have the same content while we add random salt and pepper noise in the right. We could see that even though the PSNR of the right is higher than the left one, the edge detecion result of the left is sharper.}
	\label{PSNR fail_case}
\end{figure}

Although the conventional evaluation metric such as PSNR is widely used, it isn't reasonable in some case and has drawback to present the all-sided quality of images, as shown in Fig.~\ref{PSNR fail_case}.
For clinic purpose, we focus on the sharpness of CMR. So we proposed more reasonable evaluation metric.

Learn from \cite{Kupyn2017DeblurGAN}, if an image has higher quality and sharper edge, the edge detection will return better results.
The Edge Connectivity is the most popular score to evaluate the sharpness of edges and is wildly used as evaluation metric for edge detection. In our work, we use the Edge Connectivity to evaluate the quality of correction method whose goal is to remove motion artifact and restore the sharper image from a blurred input.
A final note is that we choose the standard Sobel edge detector for our evaluation.

%https://wenku.baidu.com/view/5b735ab91a37f111f1855b1a.html
%https://www.researchgate.net/publication/314266784_Metrics_for_Edge_Detection_Methods

\section{Experiments}
In this section, we introduce our dataset used for our model and evaluate our model from both qualitative and quantitative aspects. Besides, we also adopt Structural Similarity Measure(SSIM) to evaluate the structural similarity between motion artifact input and restored output.

Because there is little open-source of existing CMR motion artifact correction methods, we can't compare our proposed method with them.
Instead of compared with other methods, we have a detail look on the loss function illustrated in \ref{loss section} and try to study the function of each term in the loss \eqref{loss function}.

\begin{figure}[htbp]
	\centerline{\includegraphics[width=9cm, height=5cm]{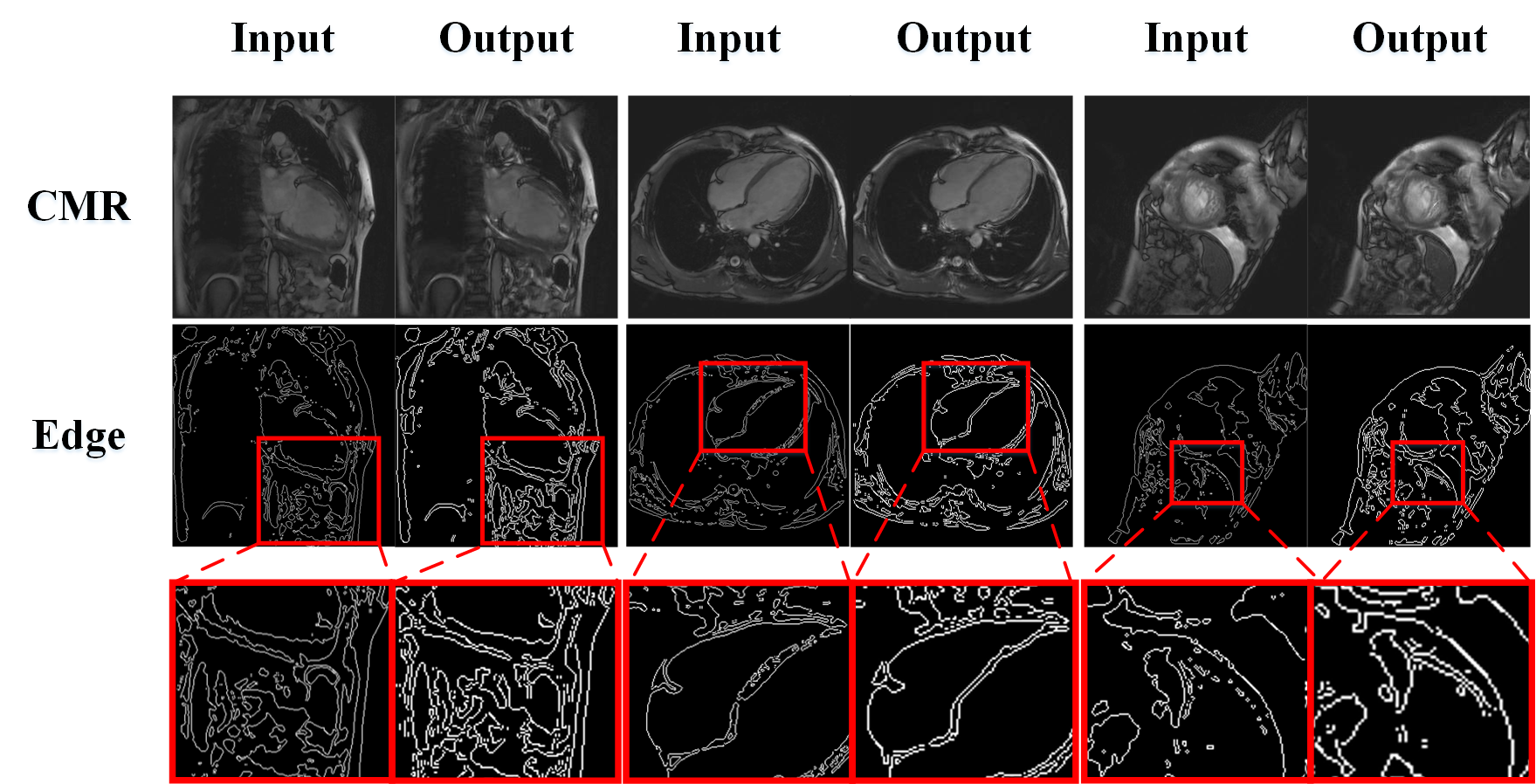}}
	\caption{Motion artifact correction model helps object edge detection. The details are highlighted in red. We can see sobel edge detections results on CMR restored by model sharper than input with motion artifact.}
	\label{help_edge_detection}
\end{figure}

As shown in Fig.~\ref{help_edge_detection}, our correction model can help edge detection results.

\subsection{Training Data}
\begin{figure}[htbp]
	\centerline{\includegraphics[width=8cm, height=4cm]{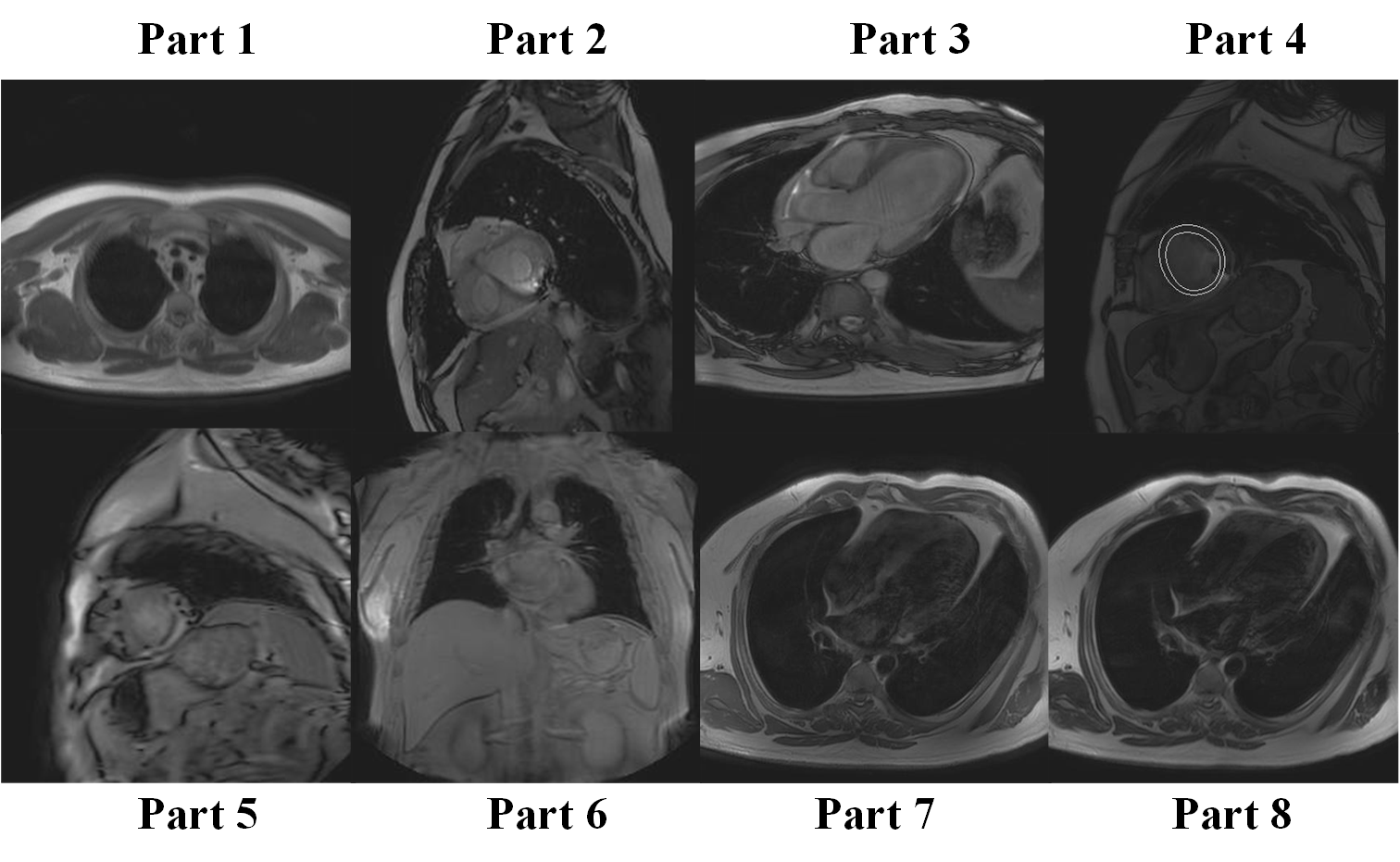}}
	\caption{Different parts of CMR}
	\label{CMR part}
\end{figure}

we collect 60 persons' CMR data from Department of MRI, Fuwai Hospital, Chinese Academy of Medical Science \& Peking Union Medical College. There are 30 patients with normal cardiac function and 30 patients with arrhythmia, each patient has 80-120 CMR images scanning from the different heart parts, details of this part are shown in Fig.~\ref{CMR part}. Totally, there are 5320 images w/o motion artifact which is used for training and 3469 images with motion artifact which are used for testing.

The raw CMR provided by FUWAI Hospital are stored in DICOM format. We transfer each DICOM to the one-channel gray image. Each image was augmented with random rigid transformation which contains random rotation, translate and zoom, then the image was cropped to 256x256 and normalized to $[0, 1]$. The $G_{\theta_G}$ take 256x256 size images as input.

\subsection{Motion artifact generation}

It is hard to get data pairs corresponding CMR with and without motion artifact from the same person for training.
Other deblur research was also confronted with this problem. We follow the idea describes by \cite{Kupyn2017DeblurGAN}.
We simulate motion artifact CMR from sharp CMR via average random several frames moving along the motion orientation of diaphragm. The trajectory is generated by Markov process.
For example, the next point of the trajectory is generated randomly based on the current point, but the moving orientation should be consistent with diaphragm.
This method allows to created a realistic image with motion artifact.

During the training phase, we use sharp images and their corresponding synthetic motion artifact images as data pairs. During the testing phase, we test the quality of our model taking the true motion artifact images as inputs.

\subsection{Ablation study on Loss function}
In this section, we have a detail look on the loss function illustrated in \ref{loss section} and try to study the function of each term in the loss \eqref{loss function}.
This can give us a fascinating insight into how the deep learning can correct CMR better.

The Tab.\ref{metric} shows the result when we sequentially adding term on the loss. The baseline refers to the ordinary pix2pix model with the content loss only and the proposed refers to our combination of content loss, edge loss and gan loss. We can see by adding edge constraint, the corrected images have high evaluation scores than only applying the content loss. By adding gan loss, we can further restore finer texture details which can help edge detect significantly.

\begin{table}[htbp]
\caption{Mean peak signal-to-noise ratio, structural similarity
measure and edge connectivity of different loss}
\begin{center}
\begin{tabular}{|c|c|c|c|}
\hline
%\textbf{Evaluation}&\multicolumn{3}{|c|}{\textbf{Models$^{\mathrm{a}}$}} \\
\textbf{Evaluation}&\multicolumn{3}{|c|}{\textbf{Models}} \\
\cline{2-4}
\textbf{Metric} & \textbf{\textit{baseline}}& \textbf{\textit{baseline+edge loss}}& \textbf{\textit{proposed}} \\
\hline
PSNR & 30.84 & 31.07 & \textbf{31.22} \\
MSSIM & 0.9197 & 0.9286 & \textbf{0.9324} \\
\hline
C/B & 0.7256 & 0.7231  & \textbf{0.7220} \\
C/A & $5.578\times10^{-3}$ & $5.577\times10^{-3}$ & \bm{$5.572\times10^{-3}$}  \\
\hline
%\multicolumn{4}{l}{$^{\mathrm{a}}$Sample of a Table footnote.}
\end{tabular}
\label{metric}
\end{center}
\end{table}

\subsection{Runtime}
We implemented all of our models using PyTorch\cite{paszke2017automatic} deep learning framework and perform the training on Texla K80 GPU. We choose Adam\cite{kingma2014adam} as a optimizer. We set learning rate initial to 0.0001 for both $G_{\theta_G}$ and $D_{\theta_D}$ and linearly decay the rate to zero after 200 epochs.  All the models are trained with batch size = 10. We utilize the code for framework released by \cite{Kupyn2017DeblurGAN}.

In addition to comparably robust and visual reality correction results, our proposed method is the first real-time CMR motion artifact correction technique. Our method just needs once forward at test time, providing a large advantage of efficiency, compared to existing complex and time-cost correction methods that need several iterations to get results. With just a single iteration, we can achieve 8.74 fps on an Tesla K80 GPU for our network to correct the motion artifact in CMR.
 A transfer from JPEG to DICOM post-process would take up to an additional 160.8 ms on the CPU. Totally, it takes only 927 ms batchsize=10 to obtain a batch of sharp CMR from blur inputs.

\section {Conclusion}
We proposed an end-to-end correction system and proposed a more reasonable evaluation metric for correction result. But whether it can be used in the clinic right now or it has been up to the clinical standard? The reality of network's output also needs to be evaluated by doctors as the clinic has a very high request for the quality and reality. A further user study that asks doctors to validate the reasonability and effectiveness of our correction system is necessary.

%\section*{Acknowledgment}
%Instead, try ``R. B. G. thanks$\ldots$''. Put sponsor acknowledgments in the unnumbered footnote on the first page \cite{IEEEexample:bluebookstandard}.

\vspace{12pt}
%\color{red}
%IEEE conference templates contain guidance text for composing and formatting conference papers. Please ensure that all template text is removed from your conference paper prior to submission to the conference. Failure to remove the template text from your paper may result in your paper not being published.

\bibliography{citation}
\end{document}